\documentclass{IEEEtran}
\IEEEoverridecommandlockouts

\usepackage{booktabs}
\usepackage{amsmath,amsfonts}
\usepackage{algorithmic}
\usepackage{algorithm}
\usepackage{array}
\usepackage[caption=false,font=scriptsize,labelfont=sf,textfont=sf]{subfig}
\usepackage{textcomp}
\usepackage{stfloats}
\usepackage{verbatim}
\usepackage{graphicx}
\usepackage{multirow}
\usepackage{amssymb}
\usepackage{amsthm}
\usepackage{mathrsfs}
\usepackage{indentfirst}
\usepackage{url}
\usepackage{makecell}
\usepackage{float}

\usepackage[normalem]{ulem}
\useunder{\uline}{\ul}{}
\usepackage{xcolor}
\usepackage[bookmarks=false,colorlinks,citecolor=black, filecolor=black, linkcolor=black,urlcolor=black]{hyperref}
\usepackage{eqparbox}
\usepackage{cleveref}
\usepackage[style=ieee, defernumbers=true]{biblatex}
\usepackage{color,url}
\usepackage{wasysym}
\usepackage{threeparttable}

\makeatletter
\let\blx@rerun@biber\relax
\makeatother

\begin{document}

\title{Interplay of Semantic Communication and Knowledge Learning}

\author{\IEEEauthorblockN{Fei Ni, Bingyan Wang, Rongpeng Li, Zhifeng Zhao and Honggang Zhang}

\thanks{  
   Fei Ni, Bingyan Wang and Rongpeng Li are with the College of Information Science and Electronic Engineering, Zhejiang University, Hangzhou 310027, China (e-mail: \{nifei; lirongpeng\}@zju.edu.cn).
   
    Zhifeng Zhao and Honggang Zhang are with Zhejiang Lab, Hangzhou, China as well as the College of Information Science and Electronic Engineering, Zhejiang University, Hangzhou 310027, China (e-mail: zhaozf@zhejianglab.com, honggangzhang@zju.edu.cn).
  }
}

\maketitle

\begin{abstract}
In the swiftly advancing realm of communication technologies, Semantic Communication (SemCom), which emphasizes knowledge understanding and processing, has emerged as a hot topic. By integrating artificial intelligence technologies, SemCom facilitates a profound understanding, analysis and transmission of communication content. In this chapter, we clarify the means of knowledge learning in SemCom with a particular focus on the utilization of Knowledge Graphs (KGs). Specifically, we first review existing efforts that combine SemCom with knowledge learning. Subsequently, we introduce a KG-enhanced SemCom system, wherein the receiver is carefully calibrated to leverage knowledge from its static knowledge base for ameliorating the decoding performance. Contingent upon this framework, we further explore potential approaches that can empower the system to operate in evolving knowledge base more effectively. Furthermore, we investigate the possibility of integration with Large Language Models (LLMs) for data augmentation, offering additional perspective into the potential implementation means of SemCom. Extensive numerical results demonstrate that the proposed framework yields superior performance on top of the KG-enhanced decoding and manifests its versatility under different scenarios.
\end{abstract}

\begin{IEEEkeywords}
Deep learning, semantic communication, knowledge graph, semantic extraction, large language model.
\end{IEEEkeywords}

\section{Introduction}
\label{sc-intro}
Advancements in Deep Learning (DL) and Natural Language Processing (NLP) have paved the way for the evolution of Semantic Communication (SemCom). As a communication paradigm that prioritizes the transmission of meaningful information over precise delivery of bits or symbols, SemCom emerges as a feasible approach to achieve high-fidelity information delivery with lower communication consumption \cite{lu2023semantics}. To boost transmission efficiency and optimize adaptability in varying channel environments, several prior studies have employed DL-based Joint Source-Channel Coding (JSCC) in the context of the SemCom system. This approach leverages an end-to-end Neural Network (NN) architecture for both transmitter and receiver, effectively conceptualizing the entire communication system as an Autoencoder (AE). Extending from this foundational structure, SemCom systems are eligible for the transmission of a wide array of data types, including but not limited to text \cite{farsad2018deep, jiang2022deep, zhou2021semantic, zhou2022adaptiveb,lu2023rethinking,lu2021reinforcement, lu2023self}, image \cite{bourtsoulatze2019deep, kurka2020deepjscc,ding2021snr, yang2023witt, wu2023transformer, lee2023deep, tong2023alternate}, speech \cite{weng2021semantic} and video \cite{jiang2022wireless}. These systems demonstrate superior performance over traditional communication systems, particularly in scenarios characterized by limited channel bandwidth and low Signal-to-Noise Ratio (SNR) environments.

However, in the abovementioned frameworks, knowledge is implicitly encapsulated within the parameters of NNs. Nonetheless, it is noteworthy that such approaches exhibit deficiencies in their capacity to thoroughly comprehend and represent knowledge. In other words, it is feasible to enhance the efficacy of SemCom systems by investigating refined knowledge representations. A prevalent approach in this endeavor is the utilization of Knowledge Graphs (KGs), which conceptualize human knowledge in the form of a graph structure. A KG comprises entities and relationships, where entities symbolize objects or concepts in the tangible world and serve as vertices within the graph. Relationships, on the other hand, represent directed edges linking two entities. By leveraging a graph structure, KGs manifest enhanced computational compatibility, interpretability, and extensibility.

Notably, the breakthroughs in Large Language Models (LLMs), exemplified by GPT series \cite{radford2018improving,radford2019language,brown2020language} and LLaMA \cite{touvron2023llama,touvron2023llama2}, have demonstrated remarkable prowess in fields of natural language understanding, generation, and reasoning. In conjunction with these advancements, Prompt Engineering (PE) emerges as a novel and versatile programming approach within LLMs, offering users a more adaptable operational framework. Going beyond traditional research boundaries, LLMs can skillfully construct and enrich KGs with their comprehension of knowledge under the instruction of appropriate prompts, thereby augmenting the usability and reliability of knowledge \cite{carta2023iterative}. Hence, it sounds appealing to utilize LLMs as an alternative data augmentation solution, particularly for extracting and contextualizing knowledge within the SemCom process, so as to enrich the depth and breadth of SemCom.

In a nutshell, this chapter aims to investigate strategies for incorporating KGs into SemCom to improve the knowledge-learning capabilities (e.g., knowledge representation and reasoning) of the SemCom system. Compared to existing studies, the main contents and innovations of this study are as follows:
\begin{itemize}
    \item A KG-enhanced SemCom system is proposed to facilitate the effective utilization of knowledge, in which a Transformer-based knowledge extractor at the receiver side is designed to find semantically related factual triples from the received noisy signal for assisting the semantic decoding.
    \item On this basis, due to the potential variations in knowledge contents, a KG evolving-based approach, which attempts to acquire possible semantic representation from received signals in a unified space, is adopted with the aid of contrastive learning \cite{chen2020simple}. It can better capture the relationship between entities and the received signals, thus boosting the flexibility and robustness of the system.
    \item In addition, this study also explores a feasible solution for data augmentation by Large Language Models. In that regard, LLMs demonstrate the feasibility of extracting the knowledge of the dataset without manual annotation through appropriate prompts.
\end{itemize}

The remainder of the chapter is organized as follows. Section \ref{sc-concepts-and-works} introduces the fundamental concepts of KGs and reviews existing works on the integration of knowledge in SemCom. Section \ref{sc-KG-semcom} describes a KG-enhanced SemCom system and presents corresponding numerical results. Furthermore, optimization strategies for the KG evolving-based SemCom system and accompanied experimental results are discussed in Section \ref{sc-KG-semcom-dynamic}. Section \ref{sc-llm} further explores a solution for data augmentation in LLMs. Finally, Section \ref{sc-conclusion} concludes the chapter.

Beforehand, we summarize the mainly used notations in Table~\ref{tab_notation}.

\begin{table}
\caption{Mainly used notations in this chapter.\label{tab_notation}}{%
\centering
\renewcommand\arraystretch{1.15}
\begin{tabular}{m{2.1cm}m{5.2cm}}
\toprule
\textbf{Notations} & \textbf{Description} \\
\midrule
$(e_p, r_{pq}, e_q)$ & Triple with head and tail entity $e_p$, $e_q$ and their relationship $r_{pq}$\\
\hline
$S_\beta(\cdot)$ & Semantic encoder with trainable parameters $\beta$ \\
\hline
$S^{-1}_\gamma(\cdot)$ & Semantic decoder with trainable parameters $\gamma$ \\
\hline
$C_\alpha(\cdot)$ & Channel encoder with trainable parameters $\alpha$ \\
\hline
$C^{-1}_\delta(\cdot)$ & Channel decoder with trainable parameters $\delta$ \\
\hline
$K_\theta(\cdot)$ & Knowledge extractor with trainable parameters $\theta$\\
\hline
$\mathbf{s}, \hat{\mathbf{s}}$ & Input and decoded sentences \\ 
\hline
$\mathbf{h}$, $\hat{\mathbf{h}}$ & Semantically encoded vector and channel decoded vector\\
\hline
$\mathbf{x}$, $\mathbf{y}$ & Transmitted and received signals \\
\hline
$\mathbf{t}$ & Indicator vector computes relevancy between triples in knowledge base and the embedding representation of $\hat{\mathbf{h}}$\\
\hline
$n_t$ & Number of triples in the knowledge base \\
\hline
$N$ & Length of sentence \\
\hline
$w$ & Weight parameter for knowledge extraction\\
\hline
$\mathcal{F}_{k}(\cdot)$, $\mathcal{F}_e(\cdot)$ & Knowledge and entity embedding process\\
\hline
$\mathcal{U}$ & Unified semantic representation space \\
\hline
$\mathbf{v}_h$ & Mapping vector of $\hat{\mathbf{h}}$ in the space $\mathcal{U}$ \\
\hline
$\mathcal{F}_{\hat{h}}(\cdot)$ & Mapping function of $\hat{\mathbf{h}}$\\
\hline
$\mathcal{D}(\cdot)$ & Generalized distance function \\
\hline
$\lambda$ & Pre-defined distance threshold \\
\hline
$\tau$ & The temperature parameter of InfoNCE \\
\hline
$\{e\}$ & The set of entities from the knowledge base \\
\hline
$\mathbf{v}_{e_i}$ & Embedding vector of $e_i$ in $\mathcal{U}$ \\
\hline
$\hat{r}_{pq}$ & Relationship computed by the relationship prediction module\\
\hline
$\{m\}$ & Extraction triples from $K_\theta(\cdot)$ or $\mathcal{U}$ \\ 
\hline
$\mathbf{k}$ & Knowledge vector obtained by $\mathcal{F}_{k}(\{m\})$\\
\hline
$\{e_s\}$ & Semantically related entity set with $\mathbf{s}$ \\
\hline
$e_+$ & Related entity from $\{e_s\}$ \\
\hline
$K$ & Number of the negative samples \\
\hline
$e_-$ & Irrelevant entity chosen from the knowledge base \\
\hline
$\mathbf{v}_{e_+}$, $\mathbf{v}_{e_-}$ & Embedding vector of $e_+$, $e_-$ as positive and negative samples \\
\bottomrule
\end{tabular}}{}
\end{table}

\section{Basic Concepts and Related Works}
\label{sc-concepts-and-works}
In this section, we commence with fundamental concepts associated with KGs, accompanied by an exploration of pertinent techniques related to KG. Subsequently, we conduct a comprehensive review of existing research regarding the integration of KGs in SemCom.

\subsection{Introduction to the KG}
\label{sc-kg}
The fundamental components of a KG encompass entities and relationships. Specifically, entities represent objects or concepts in the real world, while relationships characterize the connections between entities. From a mathematical perspective, a KG is typically represented as a graph structure, where entities form the vertices of the graph, and relationships constitute the directed edges of the graph. Each factual statement can be expressed as a triple $(e_p, r_{pq}, e_q)$, where $e_p$ and $e_q$ signify the head and tail entities, respectively, and $r_{pq}$ represents the relationship connecting the two entities.

As a computational and analytical approach, Knowledge Representation Learning (KRL), commonly referred to as knowledge embedding, endeavors to acquire low-dimensional distributional embeddings for entities and relationships within a KG. Notably, prevalent KRL methodologies can be broadly classified into two distinct categories. The first category revolves around translational distance, as exemplified by the TransE model \cite{bordes2013translating}. This approach represents entities and relationships as vectors within a predetermined mathematical space. Contrastingly, the other category leverages Deep Neural Networks (DNNs), encompassing alternative architectures such as Multilayer Perceptron (MLP), Convolutional Neural Networks (CNNs) and Graph Neural Networks (GNNs) \cite{bordes2014semantic, dettmers2018convolutional, schlichtkrull2018modeling}. These DL frameworks serve well to transform a knowledge embedding problem into a sophisticated task within the domain of NN-based approaches \cite{ji2021survey}.

\subsection{Knowledge Representation and Reasoning in SemCom}
\label{sc-kg-works}
Many existing studies on SemCom typically regard knowledge as a statistical variable underlying data acquisition procedures. In these methodologies, instead of being confined to an explicitly represented knowledge base, knowledge can be implicitly encoded as parameters of DNNs at both the transmitter and the receiver sides. Furthermore, recent literature has begun to integrate KGs into SemCom systems for a more tangible representation of knowledge. For instance, Ref. \cite{lan2021semantic} explores the viability of incorporating KGs into SemCom systems, proposing a system where KGs are adeptly merged. Targeted at both human-to-human and human-to-machine communication, the authors suggest employing KGs at the transmitter side to achieve semantic representation while utilizing KGs at the receiver side for error correction and inference. Ref. \cite{wang2022performance} focuses on enhancing communication between base stations and users, leveraging KGs as a tool for depicting semantic information. Upon obtaining the transmitted KG, receivers employ a graph-to-text model to reconstruct the information. Furthermore, the authors introduce an attention policy gradient algorithm to assess the significance of individual triples within the semantic information \cite{wang2022performance}. Similarly, Ref. \cite{zhou2022cognitive} incorporates KGs into the process of semantic encoding and decoding. This involves transforming sentences into factual triples, then encoding these into bitstreams compatible with traditional communication methods. Together with recovery and error correction procedures, the retrieved triples are then restored into meaningful semantic information utilizing a finely-tuned T5 model (a type of LLM) \cite{raffel2020exploring}. In line with this framework, the proposed scheme is further extended to implement in multi-user scenarios \cite{zhou2023cognitive}. Furthermore, Ref. \cite{jiang2022reliable} employs factual triples from KGs in their semantic encoding scheme, by measuring the semantic importance and selectively transmitting knowledge with higher semantic significance.

Besides, effective knowledge reasoning has been investigated as a by-product of SemCom. Ref. \cite{liang2022life} proposes a reasoning-based SemCom system that leverages knowledge reasoning to fill in missing or implicit entities and relationships in information. Meanwhile, in order to maintain lifelong learning, new knowledge is continuously acquired during the communication process. Ref. \cite{xiao2022reasoning} introduces a novel implicit SemCom system, which harnesses KGs for the identification and representation of latent knowledge embedded within communication content. The authors further demonstrate that the integration with federated learning contributes to collaborative reasoning across multiple users as well \cite{xiao2022imitation}. Ref. \cite{li2022cross} introduces the concept of the Cross-Modal Knowledge Graph (CKG), which establishes connections between knowledge extracted from various data sources such as video, audio and haptic signals. Subsequently, the authors investigate the viability of incorporating a CKG into a cross-modal SemCom system. Recently, Ref. \cite{jiang2023large} explores the possibility to use large-scale artificial intelligent models (i.e., LLMs) to construct the knowledge base. The authors show that the introduction of LLMs provides more precise representations of knowledge, thus reducing the need for data training and lowering computational expenses.

In a nutshell, the incorporation of KGs into contemporary research has significantly enhanced the knowledge-learning capacities of SemCom systems. Nonetheless, such integration is still at a preliminary stage and introduces extra challenges. Firstly, some studies regard factual triples as the semantic carriers \cite{zhou2022cognitive}, leading to these triples inevitably being subjected to the distortions of noisy channels during transmission. Moreover, a critical limitation arises from the inherent capacity constraints, resulting in the term ``semantic" not necessarily aligning with the ``knowledge" represented by KGs. Taking text as an illustrative example, KGs can solely depict semantics in straightforward declarative sentences, but encounter difficulties for interrogative sentences or those with subjective sentiments. Furthermore, in practice, the limited content from knowledge bases inevitably results in artificial semantic loss. Moreover, it remains under-explored to simultaneously utilize improve the decoding in SemCom by available KGs and apply received information from SemCom to evolve KGs. In other words, current research predominantly concentrates on unidirectional application of KG for semantic encoding, with a lack of emphasis on the bidirectional interaction between KGs and SemCom at the receiver side. Consequently, there emerges a strong incentive for further exploration of techniques that interplay KGs with SemCom.

\begin{figure}[t]
\centerline{\includegraphics[width=\columnwidth]{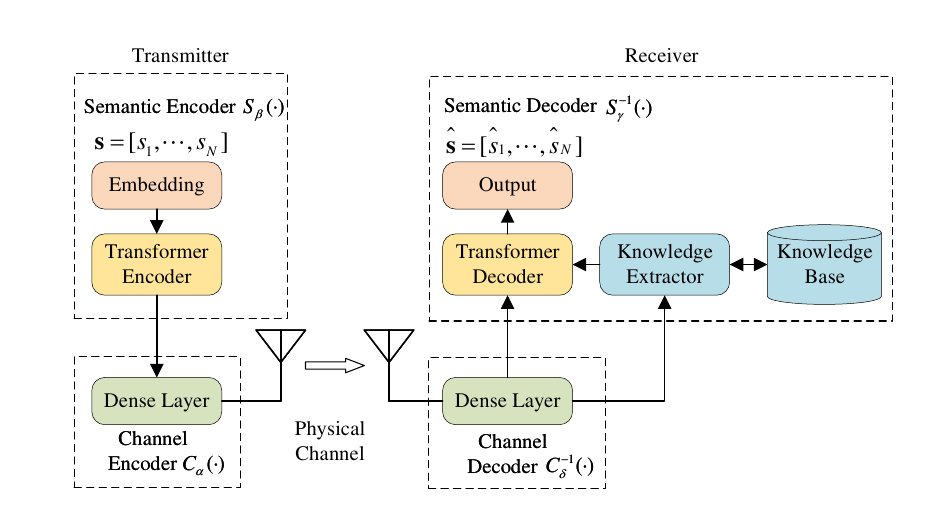}}
\vspace{-0.6cm}
\caption{The framework of the SemCom system.}
\label{fig-framework-semcom}
\end{figure}

\section{A KG-Enhanced SemCom System}
\label{sc-KG-semcom}
In this section, a KG-enhanced SemCom system is proposed, with a particular focus on the receiver side. Notably, our approach entails no requisite modifications to the DNN structure of the transmitter. Meanwhile, at the receiver side, a Transformer-based knowledge extractor is devised to extract semantically relevant factual triples from received signals and assist subsequent semantic decoding. Unlike the abovementioned studies \cite{zhou2022cognitive}, our approach positions the KG as an enhancement strategy rather than a complete carrier for semantic encoding \cite{wang2023knowledge}. Therefore, it not only enables effective knowledge-harnessing but also substantially reduces the risk of semantic loss that might arise from limitations in representational capacity or challenges in the integration of knowledge. Additionally, extensive experiments are conducted to validate the performance effectiveness and superiority.

\subsection{System Model}
\label{sc-System-Model}
The SemCom system employed in this section comprises a semantic encoder and a semantic decoder, which can be depicted in Fig.~\ref{fig-framework-semcom}. Without loss of generality, the input sentence is denoted as $\mathbf{s} = [s_1, s_2,\cdots,s_N] \in \mathbb{N}^{N}$, where $s_i$ represents the $i$-th word (i.e., token) in the sentence. In particular, the transmitter consists of two modules, that is, the semantic encoder and the channel encoder. The semantic encoder $S_\beta(\cdot)$ with trainable parameters $\beta$ extracts the semantic information in the content and represents it as a vector $\mathbf{h} \in \mathbb{R}^{N \times d_s}$, where $d_s$ is the dimension of each semantic symbol. Mathematically, 
\begin{equation}
\mathbf{h} = S_\beta(\mathbf{s}).
\label{eq_semantic_encoding}
\end{equation}

Afterwards, the channel encoder $C_\alpha(\cdot)$ with trainable parameters $\alpha$ encodes $\mathbf{h}$ into symbols that can be transmitted over the physical channel, that is,
\begin{equation}
\mathbf{x} = C_\alpha(\mathbf{h}), 
\label{eq_channel_encoding}
\end{equation}
where $\mathbf{x} \in \mathbb{C}^{N \times c}$ is the channel vector for transmission, and $c$ is the number of symbols for each token.

Given that $\mathbf{y} \in \mathbb{C}^{N \times c}$ is the received vector transmitted over the physical channel, it can be formulated as
\begin{equation}
\mathbf{y} = H\mathbf{x} + \mathbf{n}, 
\label{eq_channel_transmitting}
\end{equation}
where $H$ denotes the channel coefficient and $\mathbf{n} \sim \mathcal{N}(0, \sigma^2\mathbf{I})$ is the Additive White Gaussian Noise (AWGN).

After receiving $\mathbf{y}$, the receiver first decodes the transmitted symbols with the channel decoder $C^{-1}_\delta(\cdot)$. Hence, the decoded vector $\hat{\mathbf{h}} \in \mathbb{R}^{N \times d_s}$ can be given by
\begin{equation}
\hat{\mathbf{h}} = C^{-1}_\delta(\mathbf{y}), 
\label{eq_channel_decoding}
\end{equation}
where $\delta$ represents the trainable parameters of the channel decoder.

Notably, SemCom implicitly relies on some prior knowledge between the transmitter and receiver for the joint training process. However, different from such prior knowledge, the knowledge base in our model refers to some factual triples and can be located at the receiver side. To exploit the knowledge base, a knowledge extractor is further applied at the receiver side to extract and integrate relevant knowledge from the received signal to yield the aggregated knowledge $\textbf{k}$. In particular, the knowledge extraction and aggregation process can be formulated as
\begin{equation}
\mathbf{k} = K_\theta(\hat{\mathbf{h}}), 
\label{eq_knowledge_extraction}
\end{equation}
where $K_\theta(\cdot)$ represents the knowledge extractor with trainable parameters $\theta$. 

Eventually, the knowledge enhanced semantic decoder $S^{-1}_\gamma(\cdot)$ with trainable parameters $\gamma$ leverages the channel decoded vector $\hat{\mathbf{h}}$ and the extracted knowledge vector $\mathbf{k}$ to obtain the restored message $\hat{\mathbf{s}} = [\hat{s}_1, \hat{s}_2,...,\hat{s}_N]$, which can be expressed as
\begin{equation}
\hat{\mathbf{s}} = S^{-1}_\gamma(\hat{\mathbf{h}} \  || \  \mathbf{k}), 
\label{eq_semantic_decoding}
\end{equation}
where $||$ indicates a concatenation operator. 

Generally, the accuracy of SemCom is determined by the semantic similarity between the transmitted and received contents. In order to minimize the semantic errors between $\mathbf{s}$ and $\hat{\mathbf{s}}$, the loss function $\mathcal{L}_\text{model}$ taking account of the Cross Entropy (CE) of the two vectors can be formulated as
\begin{equation}
\mathcal{L}_\text{model} = - \sum_{i=1}^{N}\left(q(s_i)\log{p(\hat{s}_i)}\right),
\label{eq_semantic_similarity}
\end{equation}
where $q(s_i)$ is the one-hot representation of $s_i$ with $s_i \in \mathbf{s}$ and $p(\hat{s}_i)$ is the predicted probability of the $i$-th word. 

Instead of drawing upon traditional communication modules for physical-layer transmission, most existing studies have chosen to utilize end-to-end DNNs to accomplish the whole communication process. Typically, the semantic encoders and decoders are based on Transformer \cite{vaswani2017attention}. Meanwhile, the channel encoding and decoding part can be viewed as an AE implemented by fully connected layers. The whole SemCom process is then reformulated as a sequence-to-sequence problem. On this basis, we primarily focus on developing appropriate implementation means of the KG-enhanced SemCom receiver, i.e., the knowledge extractor in Eq.~\eqref{eq_knowledge_extraction} and the knowledge-enhanced decoder in Eq.~\eqref{eq_semantic_decoding}, thereby minimizing $\mathcal{L}_\text{model}$ in Eq.~\eqref{eq_semantic_similarity}.

\begin{figure}[t]
\centerline{\includegraphics[width=\linewidth]{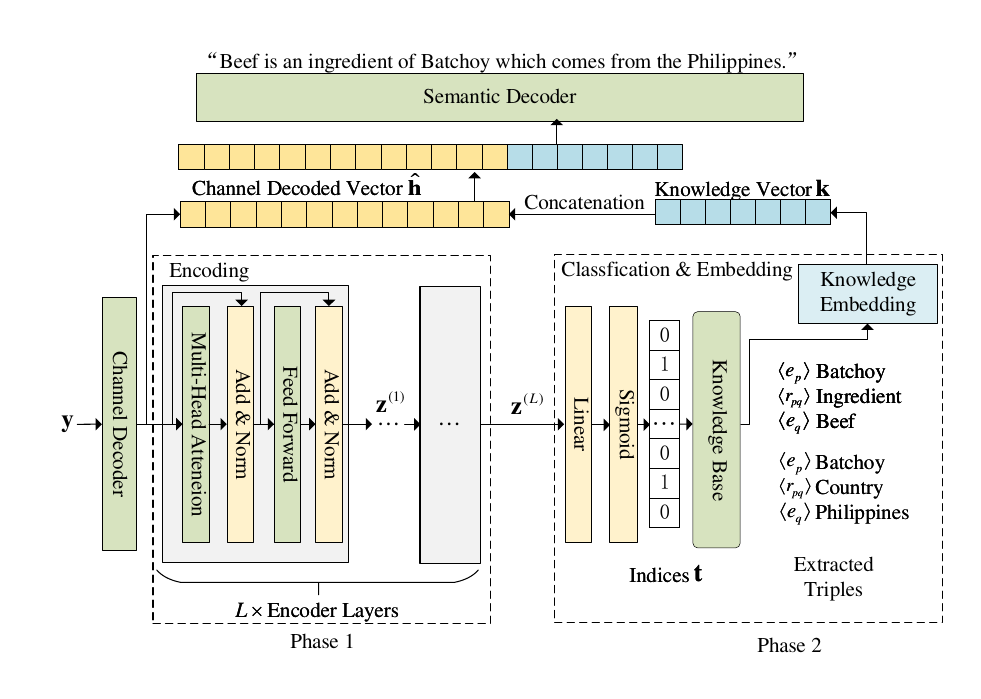}}
\vspace{-0.8cm}
\caption{The KG-enhanced semantic decoder.}
\label{fig-KG-decoder}
\end{figure}

\subsection{The Design of the KG-enhanced SemCom Receiver}
\label{sc-extractor}
In this part, we begin with the implementation details of the knowledge extractor. The whole knowledge extraction process, as illustrated in Fig.~\ref{fig-KG-decoder}, can be divided into two phases. The first phase initially involves an embedding task, by utilizing Transformer encoders to derive a representation of the decoded vector. Subsequently, in the second phase, it turns to identify all triples corresponding with the representation via a multi-label classifier. Following the identification, these triples are then compacted into a condensed format, which plays a crucial role in facilitating the final decoding.

In particular, in order to extract the semantic representation, we adopt a model composed of a stack of $L$ identical Transformer encoder layers, each of them consisting of a multi-head attention mechanism, as well as some feed-forward and normalization sublayers \cite{vaswani2017attention}. Without loss of generality, assuming that $\mathbf{z}^{(l-1)}$ is the output of the $({l-1})$-th encoder layer, where $\mathbf{z}^{(0)}$ is equivalent to $\hat{\mathbf{h}}$ for the input layer, the self-attention mechanism of the $l$-th layer can be represented as
\begin{equation}
\text{Attention}(\mathbf{z}^{(l-1)}) = \text{softmax}(\frac{\mathbf{Q}^{(l)}\left(\mathbf{K}^{(l)}\right)^\intercal}{\sqrt{d_k}})\mathbf{V}^{(l)},
\label{eq_attention}
\end{equation}
where $\mathbf{Q}^{(l)} = \mathbf{z}^{(l-1)} \mathbf{W}_Q^{(l)} $, $\mathbf{K}^{(l)} = \mathbf{z}^{(l-1)} \mathbf{W}_K^{(l)} $, $\mathbf{V}^{(l)} = \mathbf{z}^{(l-1)} \mathbf{W}_V^{(l)} $. $\mathbf{W}_Q^{(l)}$, $\mathbf{W}_K^{(l)}$ and $\mathbf{W}_V^{(l)}$ are the projection matrices of the $l$-th layer, and $d_k$ is the model dimension. Furthermore, $\mathbf{z}^{(l-1)}$ is added to the calculated $\text{Attention}(\mathbf{z}^{(l-1)})$ via a normalized residual connection to obtain the intermediate output
\begin{equation}
\mathbf{a}^{(l)} = \text{LayerNorm}(\text{Attention}(\mathbf{z}^{(l-1)}) + \mathbf{z}^{(l-1)}),
\label{eq_add_and_norm}
\end{equation}
where $\text{LayerNorm}(\cdot)$ denotes a layer normalization operation. Afterwards, a feed-forward network is involved as $\text{FFN}(\mathbf{a}^{(l)}) = \text{max}(0, \mathbf{a}^{(l)} \mathbf{W}_\text{F1}^{(l)} + \mathbf{b}_\text{F1}^{(l)}) \mathbf{W}_\text{F2}^{(l)} + \mathbf{b}_\text{F2}^{(l)}$, where $\mathbf{W}_\text{F1}^{(l)}$, $\mathbf{W}_\text{F2}^{(l)}$, $\mathbf{b}_\text{F1}^{(l)}$ and $\mathbf{b}_\text{F2}^{(l)}$ are parameters of the feed-forward layer in the $l$-th encoder block. Next, we adopt a residual connection and a layer normalization that can be formulated as
\begin{equation}
\mathbf{z}^{(l)} = \text{LayerNorm}(\text{FFN}(\mathbf{a}^{(l)}) + \mathbf{a}^{(l)}).
\label{eq_add_and_norm2}
\end{equation}

After $L$ recursive encoding layers, the embedding representation $\mathbf{z}^{(L)}$ of the channel decoded vector $\hat{\mathbf{h}}$ is obtained. Then an indicator vector $\mathbf{t}$ is derived contingent on a multi-label classifier which computes the relevancy between triples and the representation, that is,
\begin{equation}
\mathbf{t} = \text{sigmoid}(\mathbf{z}^{(L)} \mathbf{W}_t  + \mathbf{b}_t),
\label{eq_prediction}
\end{equation}
where $\mathbf{t} = [\hat{t}_1, \cdots, \hat{t}_{n_t}] \in \mathbb{R}^{n_t}, \hat{t}_i \in [0, 1]$ for $i \in \{1, \cdots, n_t\}$. Besides, $n_t$ denotes the number of triples in the knowledge base, while $\mathbf{W}_t$ and $\mathbf{b}_t$ are parameters of the classifier. If $\hat{t}_i \ge 0.5$, the triple $m_i$ corresponding to index $i$ is predicted to be relevant to the received content. 

Ultimately, the set of relevant factual triples $\{m\}$ predicted by the model are embedded into the knowledge vector $\mathbf{k} = \mathcal{F}_{k}(\{m\})$, where the embedding process is abstractly represented as $\mathcal{F}_{k}(\cdot)$. In particular, rather than compute the embedding of the entity and relationship separately, we choose to integrate the triples into the same compressed format. Subsequently, as in Eq.~\eqref{eq_semantic_decoding}, the knowledge vector is concatenated with the decoding vector and fed into the semantic decoder. Therefore, the extracted knowledge $\mathbf{k}$ from $\hat{\mathbf{h}}$ with the assistance of KGs is capable to ameliorate the semantic decoding process. The procedure of the KG-enhanced SemCom system is illustrated in Algorithm ~\ref{alg-KG-reciver}.

\begin{algorithm}[tbp]
\caption{The SemCom Process with the KG-Enhanced Receiver.} 
\label{alg-KG-reciver}
\begin{algorithmic}[1]
\renewcommand{\algorithmicrequire}{\textbf{Initialization:}}
\REQUIRE The parameters of the semantic encoder $S_\beta(\cdot)$, channel encoder $C_\alpha(\cdot)$, semantic decoder $S^{-1}_\gamma(\cdot)$, channel decoder $C^{-1}_\delta(\cdot)$ and knowledge extractor $K_\theta(\cdot)$. \\
\renewcommand{\algorithmicrequire}{\textbf{Input:}}
\REQUIRE The tokenized sentence $\mathbf{s}$. \\
\renewcommand{\algorithmicrequire}{\textbf{Output:}}
\REQUIRE The restored sentence $\hat{\mathbf{s}}$.
\vspace{.1cm}\\
\textit{Transmitter:}
\STATE Semantic encoding: $\mathbf{h} \gets (S_\beta(\mathbf{s}))$.
\STATE Channel encoding: $\mathbf{x} \gets (C_\alpha(\mathbf{h}))$.
\STATE Transmit $\mathbf{x}$ over the physical channel: $\mathbf{y} \gets \mathbf{Hx} + \mathbf{n}$.
\vspace{.1cm}\\
\textit{Receiver:}
\STATE Channel decoding: $\hat{\mathbf{h}} \gets C^{-1}_\delta(\mathbf{y})$.
\STATE Knowledge extraction $K_\theta(\cdot)$:
\STATE \quad Compute the embedding representation $\mathbf{z}^{(L)}$.
\STATE \quad $\mathbf{t} \gets \text{sigmoid}(\mathbf{z}^{(L)} \mathbf{W}_t  + \mathbf{b}_t)$.
\STATE \quad Find the triples $\{m\}$ where $\hat{t}_i \ge 0.5$.
\STATE \quad Knowledge embedding: $\mathbf{k} \gets \mathcal{F}_{k}(\{m\})$.
\STATE Semantic decoding: $\hat{\mathbf{s}} \gets S^{-1}_\gamma(\hat{\mathbf{h}} \  || \  \mathbf{k})$.
\end{algorithmic} 
\end{algorithm}

\subsection{The Training Methodology}
\label{sc-training}
The training of the knowledge extractor $K_\theta(\cdot)$ heavily relies on an end-to-end SemCom model. On this basis, the sentences are first transmitted by the transmitter via the channel and then the channel decoded vectors are fed into the knowledge extractor. Afterwards, the knowledge extractor is trained by gradient descent while the parameters of the transmitter are frozen. Since the number of negative labels is much more than that of positive labels, the loss function adopts the weighted Binary Cross Entropy (BCE), which can be represented as
\begin{equation}
\mathcal{L}_\text{knowledge} = \sum_{i=1}^{n_t} -w_i [t_i \cdot \log{\hat{t}_i} + (1 - t_i) \cdot \log(1 - \hat{t}_i)]. \label{eq_loss_function}
\end{equation}
Here $t_i \in \{0, 1\}$ represents the training label, and $\hat{t}_i$ is the prediction output in Eq.~\eqref{eq_prediction}. $w_i$ is the weight of the $i$-th position, related to the hyperparameter $w$. Specifically, $w_i = w$ when $t_i = 0$, otherwise $w_i = 1 - w$. Increasing $w$ can result in a more sensitive extractor, but it also brings an increase in the false positive rate.

Same as the Transformer encoder, the training complexity of a knowledge extractor is $O(LN^2 \cdot d_k)$. Notably, the knowledge extractor is not limited to the conventional Transformer structure, but can also be applied to different Transformer variants, such as Universal Transformer (UT) \cite{dehghani2018universal}. With the self-attention mechanism, the extracted factual triples can provide additional prior knowledge to the semantic decoder and therefore improve the performance of semantic decoding. Typically, the knowledge vector is concatenated to the received message, rather than being merged into the encoded vector of the input sentence as done by previous works \cite{zhou2022cognitive}. Therefore, our works ensures that when the extractor is of little avail (e.g., no relevant knowledge found by the knowledge extractor), the system can function normally by using a standard encoder-decoder Transformer structure, while avoiding possible semantic losses introduced by the knowledge extraction procedure.

\begin{table}
\caption{Experimental settings.\label{tab-KG-decoder-parameter}}
\centering
\begin{tabular}{cc}
\toprule
Parameter & value \\
\midrule
Train dataset size & $24,467$ \\ 
Test dataset size  & $2,734$  \\
Weight parameter $w$  & $0.02$  \\
DNN Optimizer & Adam \\
Batch size & $32$ \\
Model dimension & $128$ \\
Learning rate & $10^{-4}$ \\
Channel vector dimension & $16$ \\
No. of multi-heads & $8$ \\
\bottomrule
\end{tabular}
\end{table}

\subsection{Simulation Results}
\label{sc-results}
\paragraph{Dataset and Parameter Settings}
\label{paragraph-dataset}
The dataset used in the numerical experiment is based on WebNLG v3.0 \cite{gardent2017creating}, which consists of data-text pairs. For each data-text pair, the data is a set of triples extracted from DBpedia while the text corresponds to the verbalization of these triples. In this numerical experiment, the weight parameter $w$ is set to $0.02$, while the learning rate is set to $10^{-4}$. Moreover, we set the dimension of the dense layer as $128 \times 16$, and adopt 8-head attention in transformer layers. The detailed settings of the proposed system are shown in Table~\ref{tab-KG-decoder-parameter}. We train the models based on both the classical Transformer and UT \cite{zhou2021semantic}. Besides, we adopt two metrics to evaluate their performance, that is, 1-gram Bilingual Evaluation Understudy (BLEU) \cite{papineni2002bleu} score for measuring word-level accuracy and Sentence-BERT \cite{reimers2019sentence} score for measuring semantic similarity. Notably, as a Siamese Bert-network model that generates fixed-length vector representations for sentences, Sentence-BERT produces a score in terms of the cosine similarity of embedded vectors.

\begin{figure*}[t]
\centering
\includegraphics[width=0.8\textwidth]{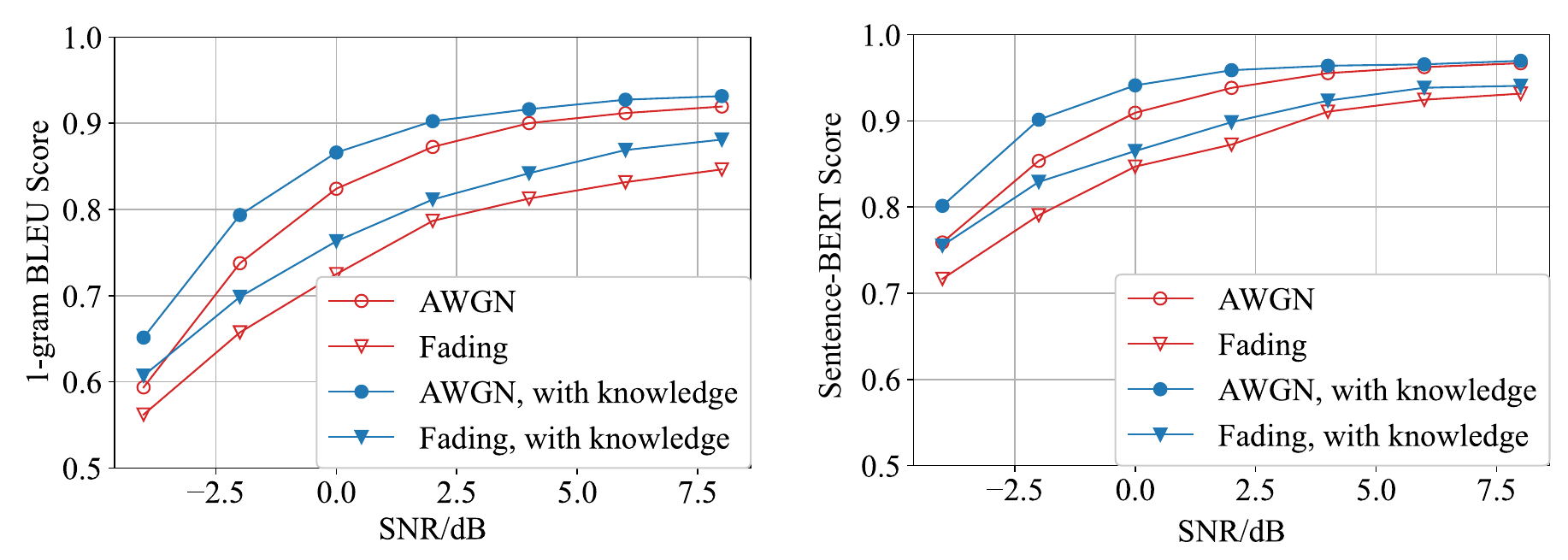}
\vspace{-0.3cm}
\caption{The BLEU and Sentence-BERT score versus SNR for the KG-enahnced SemCom system based on Transformer.
\label{fig-KG-decoder-results-transformer}}
\end{figure*}

\begin{figure*}[t]
\centering
\includegraphics[width=0.8\textwidth]{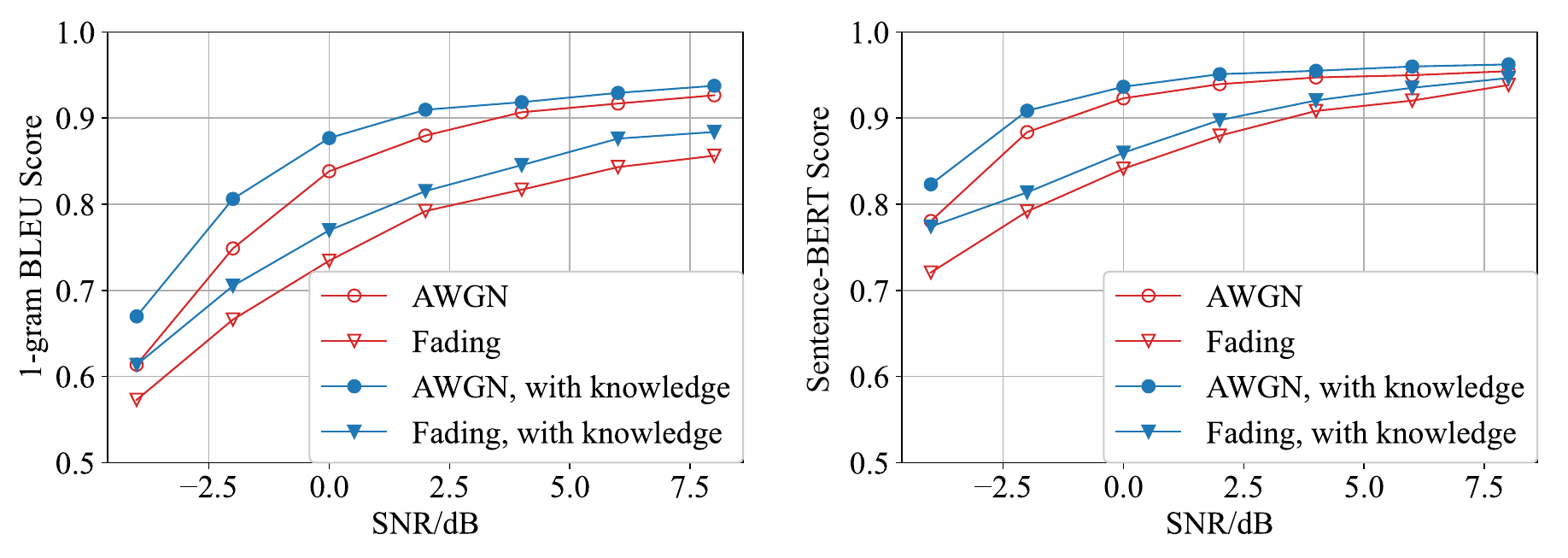}
\vspace{-0.3cm}
\caption{The BLEU and Sentence-BERT score versus SNR for the KG-enahnced SemCom system based on UT.
\label{fig-KG-decoder-results-UT}}
\end{figure*}

\begin{table}[t]
\caption{Example of semantic decoding process based on knowledge enhancement.\label{tab-KG-decoder-exmaple-decoding}}
\centering
\begin{tabular}{c|c}
\toprule
Enter sentence & \makecell[l]{Ayam penyet contains fried chicken and \\ originates from Singapore. It can also be \\ found in Java where the banyumasan \\ people are one of the ethnic groups.}\\ \midrule
\makecell[c]{Related knowledge in \\ the knowledge base} & \makecell[l]{(Ayam\_penyet, ingredient, Fried\_chicken) \\ (Java, ethnicGroup, Banyumasan\_people) \\ (Ayam \_penyet, region, Singapore) \\ (Ayam\_penyet, country, Java)} \\ \midrule
\makecell[c]{Decoded results without \\ knowledge enhancement} & \makecell[l]{Ayam penyet contains fried \textbf{tomatoes} and \\ originates from Singapore. It can also be \\ found in \textbf{Japan} where the banyumasan \\ people are one of the ethnic groups.}\\ \midrule
\makecell[c]{knowledge-enhanced \\ decoding results} & \makecell[l]{Ayam penyet contains fried chicken and \\ originates from Singapore. It can also be \\ found in Java where the banyumasan \\ people are one of the ethnic groups.}\\ \bottomrule
\end{tabular}
\end{table}

\paragraph{Numerical Results}
\label{paragraph-results}
Fig.~\ref{fig-KG-decoder-results-transformer} shows the BLEU and Sentence-BERT score versus SNR for the KG-enhanced SemCom system based on Transformer, respectively. It can be observed that the assistance of the knowledge extractor could significantly contribute to improving the performance. In particular, regardless of the channel type, the knowledge extractor can always bring more than $5\%$ improvement in BLEU under low SNRs. For the Sentence-BERT score, the KG-enhanced receiver also shows a similar performance improvement. The results demonstrate that the proposed scheme can improve the comprehension of semantics at the receiver side. On the other hand, Fig.~\ref{fig-KG-decoder-results-UT} manifests the performance of the system based on UT under both the BLEU and Sentence-BERT score, and a similar performance superiority could also be expected. In order to demonstrate the effect of knowledge enhancement more intuitively, Table~\ref{tab-KG-decoder-exmaple-decoding} shows a case in the semantic decoding process. In particular, the original text characterizes a food item, encompassing related details such as its ingredients and the country of origin. Due to the annoying channel noise, the received signal may encounter some semantic errors. Although traditional semantic decoders are able to correct grammatical errors, other errors in semantics and knowledge may be ignored. Instead, the knowledge extractor can locate related knowledge from the knowledge base, so as to assist the semantic decoder for the output of the correct result.

\begin{figure}[!tb]
\centering
\includegraphics[width=0.4\textwidth]{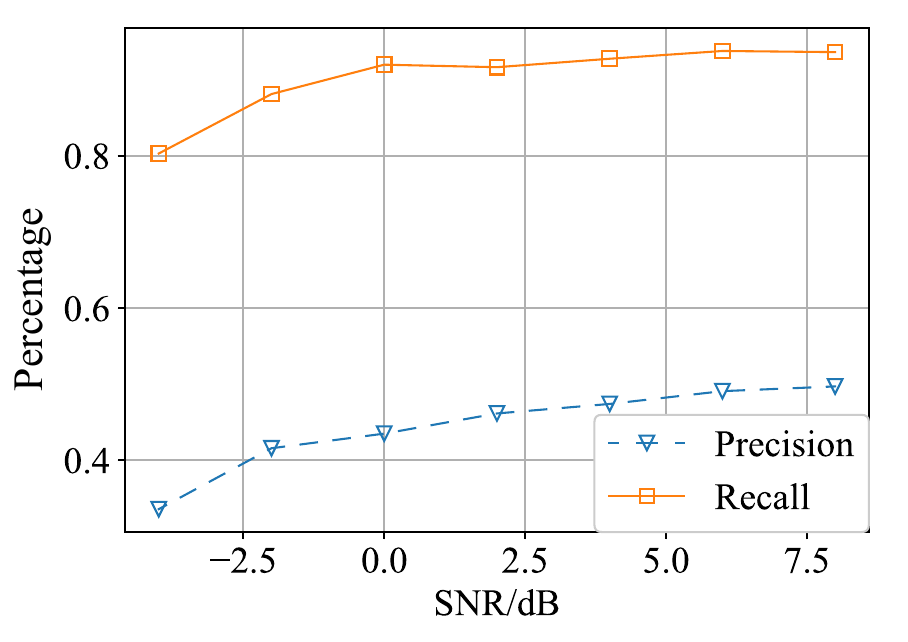}
\vspace{-0.6cm}
\caption{The precision and recall rate versus SNR for the KG-enhanced SemCom system based on Transformer.
\label{fig-KG-decoder-results-precision}}
\end{figure}

\begin{figure}[!tb]
\centering
\includegraphics[width=0.4\textwidth]{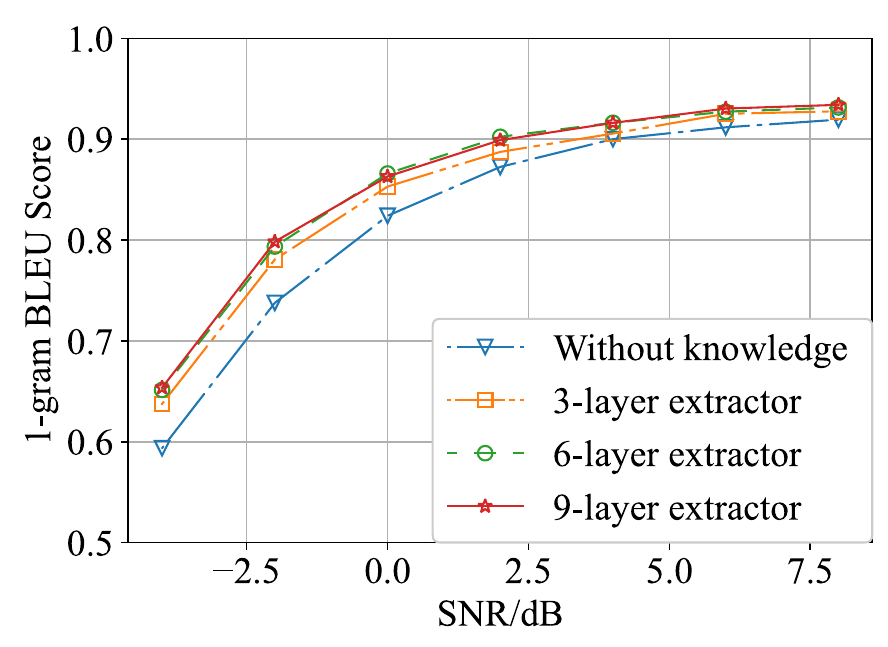}
\vspace{-0.5cm}
\caption{The comparison of the BLEU score for knowledge extractor with different numbers of Transformer encoder layers.
\label{fig-KG-decoder-results-encoder-layer}}
\end{figure}

On the other hand, we test the precision and recall rate versus SNR for the KG-enhanced SemCom system based on Transformer. As shown in Fig.~\ref{fig-KG-decoder-results-precision}, the KG-enhanced SemCom system can obtain a recall rate of over $90\%$. However, the received content may be polluted by noise, resulting in an increase of false positives and leading to a large gap between precision and recall. 

Intuitively, the number of encoder layers in the knowledge extractor may also affect the performance of the system. Therefore, we additionally implement the knowledge extractor with different number of Transformer encoder layers and present the BLEU performance comparison in Fig.~\ref{fig-KG-decoder-results-encoder-layer}. It can be observed that the $6$-layer model performs slightly better than the $3$-layer model. However, the performance remains almost unchanged when it increases to $9$ layers. Furthermore, in addition to utilize the fixed model trained at a certain SNR, it is also possible to leverage several SNR-specific models, each corresponding to a specific SNR. Table~\ref{tab-KG-decoder-snr-fixed-specific} demonstrates the BLEU performance comparison between the fixed model trained at an SNR of $0$ dB and different SNR-specific models. It can be observed that compared to the fixed model, the SNR-specific model could yield superior performance improvements at the corresponding SNR.

\begin{table}[t]
\caption{The comparison of the BLEU score between a fixed extractor model and SNR-specific models.\label{tab-KG-decoder-snr-fixed-specific}}
\centering
\begin{tabular}{ccc}
\toprule
SNR/dB & Fixed & SNR-specific  \\
\midrule
-4 &  0.6514 &  0.6718 \\ 
-2 &  0.7936 &  0.8126 \\ 
0  &  0.8661 &  0.8661 \\ 
2  &  0.9025 &  0.9134 \\ 
4  &  0.9164 &  0.9201 \\
\bottomrule
\end{tabular}
\end{table}

\section{A KG Evolving-Based SemCom System}
\label{sc-KG-semcom-dynamic}
After exploring a KG-enhanced SemCom system in Section \ref{sc-KG-semcom}, in this section, we further investigate potential means to empower the system in order to more effectively operate in evolving knowledge scenarios.

\subsection{System Overview}
The approach in Section \ref{sc-KG-semcom} manifests the performance of a static KG-enhanced means that significantly depends on the knowledge base with a limited capacity. Thus, it mandates the need for continually retraining the classifier to capture potential updates in the knowledge base, as in practical applications, complex knowledge-processing activities are often accompanied by frequent and subtle variations in knowledge content over time. Hence, in order to maintain the essential flexibility and adaptability of the knowledge extractor, this section proposes a KG evolving-based SemCom system on top of a dynamic knowledge base at the receiver side, by incorporating the concept of unified semantic representation space. Notably, the framework of KG evolving-based SemCom is the same as KG-enhanced SemCom depicted in Section \ref{sc-System-Model}, with a particular focus on the design of the receiver on account of a dynamic knowledge base.

\subsection{The Design of the KG evolving-based SemCom Receiver}
\label{sc-extractor-dynamic}

\begin{figure*}[t]
\centering
\includegraphics[width=0.6\textwidth]{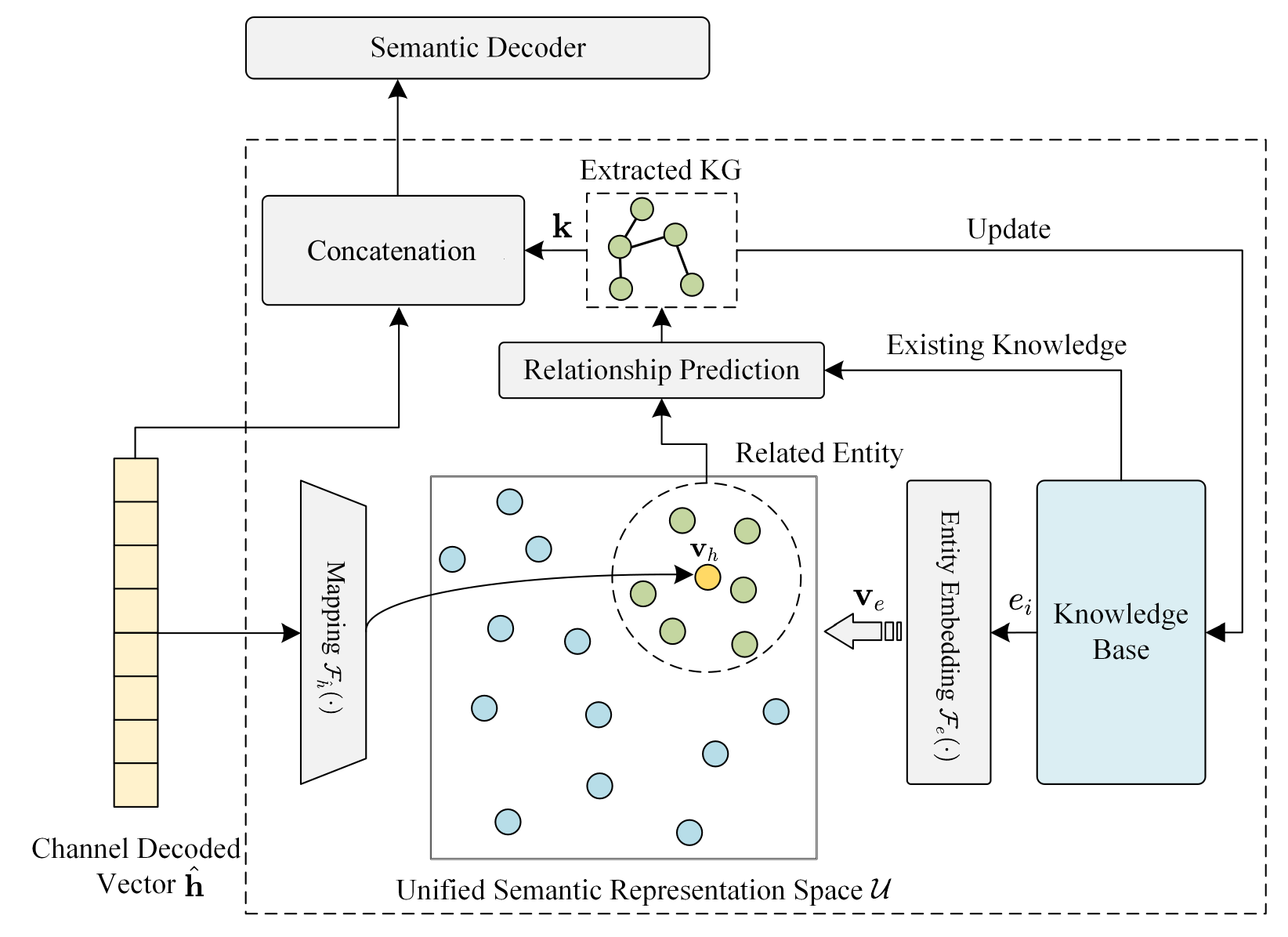}
\caption{The structure of the KG evolving-based SemCom receiver.}  %The SemCom receiver based on the evolving knowledge base
\label{fig-dynamic-KG-decoder}
\end{figure*}

The structure of the KG evolving-based SemCom receiver is depicted in Fig.~\ref{fig-dynamic-KG-decoder}. Similar to the framework proposed in Section \ref{sc-KG-semcom}, the receiver contains a knowledge base in which the knowledge is organized and stored in the form of factual triples. However, different from Section \ref{sc-KG-semcom}, a unified semantic representation space $\mathcal{U} \in \mathbb{R}^d$ is added here, where $d$ denotes the space dimension. Notably, the unified semantic representation space implies that each entity $e_i$ belongs to the entity set $\{e\}$ from the knowledge base has a corresponding embedding vector in the space $\mathcal{U}$, which can be formulated as
\begin{equation}
\mathbf{v}_{e_i} = \mathcal{F}_{e}(e_i),
\label{eq_e_mapping}
\end{equation}
where $\mathcal{F}_{e}(\cdot)$ is the abstraction of the entire embedding process and can be implemented by DNNs. 

After obtaining the channel decoded vector $\hat{\mathbf{h}}$ as in Eq.~\eqref{eq_channel_decoding}, the receiver maps $\hat{\mathbf{h}}$ to the unified semantic representation vector $\mathbf {v}_h$ according to a DNN-induced mapping function $\mathcal{F}_{\hat{h}}(\cdot)$, that is,
\begin{equation}
\mathbf{v}_h = \mathcal{F}_{\hat{h}}(\hat{\mathbf{h}}).
\label{eq_s_mapping}
\end{equation}
Taking the linear mapping as an example of $\mathcal{F}_{\hat{h}}(\cdot)$, Eq.~\eqref{eq_s_mapping} can be re-written as $\mathbf{v}_h = \mathbf{W}_h \cdot \hat{\mathbf{h}} + \mathbf{b}_h$, where $\mathbf{W}_h$ and $\mathbf{b}_h$ are trainable parameters of the DNN. In addition to linear mapping, more complex DNN structures, such as Transformer, can also be used to fully exploit the semantic features of the signal.

By leveraging the mapping, the receiver tries to find all suitable embedding vectors (and accompanied entities) $\mathbf{v}_{e_i}, e_i \in \{e\}$ whose distances from $\mathbf{v}_h$ are within a preset threshold $\lambda$, that is, 
\begin{equation}
\mathcal{D}(\mathbf{v}_h, \mathbf{v}_{e_i}) \le \lambda.
\label{eq_distance}
\end{equation}
Here $\mathcal{D}(\cdot)$ represents a generalized distance function and can be calibrated according to the training method of the unified semantic space. Common forms of the distance function include Euclidean distance $\mathcal{D}_\text{euclid}( \mathbf{x}, \mathbf{y}) = \sqrt{\lVert\mathbf{x} - \mathbf{y}\lVert^2}$, cosine distance $\mathcal{D}_\text{cosine}(\mathbf{x}, \mathbf{y}) = 1 - \frac{\mathbf{x} \cdot \mathbf{y}}{\lVert \mathbf{x} \rVert \cdot \lVert \mathbf{y} \rVert}$, etc. All entities satisfying Eq.~\eqref{eq_distance} are added to a set $\{e_s\}$.

\begin{figure*}[t]
\centering
\includegraphics[width=0.6\textwidth]{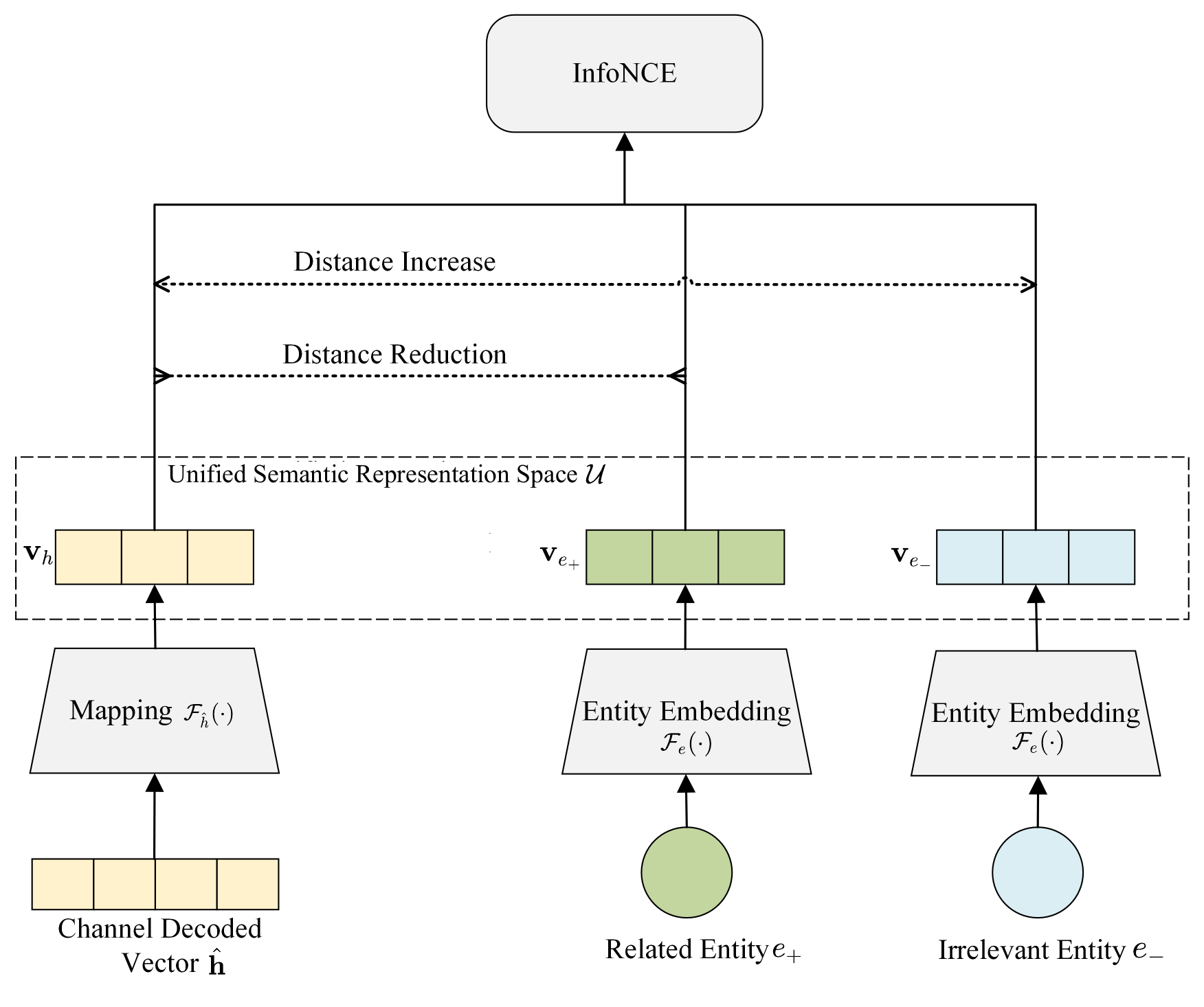}
\vspace{-0.5cm}
\caption{The schematic diagram of the training process of unified semantic representation.}
\label{fig-dynamic-KG-decoder-contrastive}
\end{figure*}

The identification of the entity set $\{e_s\}$ from the knowledge base lays the very foundation for extracting factual triples $\{m\}$. To elaborate, for each entity pair $e_p$ and $e_q$ within the set $\{e_s\}$, the presence of a pre-existing triple $(e_p, r_{pq}, e_q)$ in the knowledge base leads to its incorporation into the set of extraction results $\{m\}$. On the other hand, for instances where such a triple is absent, a relationship prediction module, which can be independently trained as a classification model with the CE loss function, is invoked to ascertain the potential existence of a relational $\hat{r}_{pq}$ between $e_p$ and $e_q$. In cases where $\hat{r}_{pq}$ is predicted, this newly identified triple $(e_p, \hat{r}_{pq}, e_q)$ is then appended to the set of extraction results. Moreover, it can be updated to the existing knowledge base, thus leading to evolving KGs. Subsequently, subject to a knowledge embedding procedure, all extracted triples $\{m\}$ contribute to generating the knowledge vector $\mathbf{k}$. Consistent with the description in Section \ref{sc-KG-semcom}, this vector $\mathbf{k}$ is concatenated with the channel decoded vector $\hat{\mathbf{h}}$. Meanwhile, the missing factual triples are further incorporated in the knowledge base, so as to facilitate subsequent decoding. The procedure of the KG evolving-based SemCom system is summarized in Algorithm ~\ref{alg-optimized-KG-reciver}.

\begin{algorithm}[t]
\caption{The SemCom Process with the KG Evolving-based Receiver.} 
\label{alg-optimized-KG-reciver}
\begin{algorithmic}[1]
\renewcommand{\algorithmicrequire}{\textbf{Initialization:}}
\REQUIRE The parameters of the semantic encoder $S_\beta(\cdot)$, channel encoder $C_\alpha(\cdot)$, semantic decoder $S^{-1}_\gamma(\cdot)$, channel decoder $C^{-1}_\delta(\cdot)$ and the unified semantic representation space $\mathcal{U}$. \\
\renewcommand{\algorithmicrequire}{\textbf{Input:}}
\REQUIRE The tokenized sentence $\mathbf{s}$. \\
\renewcommand{\algorithmicrequire}{\textbf{Output:}}
\REQUIRE The restored sentence $\hat{\mathbf{s}}$.
\vspace{.1cm}\\
\textit{Transmitter:}
\STATE Semantic encoding: $\mathbf{h} \gets (S_\beta(\mathbf{s}))$.
\STATE Channel encoding: $\mathbf{x} \gets (C_\alpha(\mathbf{h}))$.
\STATE Transmit $\mathbf{x}$ over the physical channel: $\mathbf{y} \gets \mathbf{Hx} + \mathbf{n}$.
\vspace{.1cm}\\
\textit{Receiver:}
\STATE Channel decoding: $\hat{\mathbf{h}} \gets C^{-1}_\delta(\mathbf{y})$.
\STATE Compute unified semantic representation vector $\mathbf {v}_h$ with Eq.~\eqref{eq_s_mapping}.
\STATE Find all suitable embedding vectors $\mathbf{v}_{e_i}, e_i \in \{e\}$ with Eq.~\eqref{eq_distance} and add the corresponding entities into the set $\{e_s\}$.
\FOR{each entity pair $e_p$ and $e_q$ in $\{e_s\}$}
\IF{$(e_p, r_{pq}, e_q)$ exists in the knowledge base}
\STATE Integrate them into the extraction triples $\{m\}$. 
\ELSE
\STATE Obtain the relationship $\hat{r}_{pq}$ between $e_p$ and $e_q$ by the relationship prediction module.
\STATE Add the corresponding triple $(e_p, \hat{r}_{pq}, e_q)$ to $\{m\}$.
\STATE Update $(e_p, \hat{r}_{pq}, e_q)$ to the existing knowledge base.
\ENDIF
\ENDFOR
\STATE Knowledge embedding: $\mathbf{k} \gets \mathcal{F}_{k}(\{m\})$.
\STATE Semantic decoding: $\hat{\mathbf{s}} \gets S^{-1}_\gamma(\hat{\mathbf{h}} \  || \  \mathbf{k})$.
\end{algorithmic} 
\end{algorithm}

\subsection{The Training Methodology}
\label{sc-training-dynamic}
As illustrated in Fig.~\ref{fig-dynamic-KG-decoder-contrastive}, we introduce contrastive learning \cite{chen2020simple} to train $\mathcal{F}_{e}(\cdot)$ and $\mathcal{F}_{\hat{h}}(\cdot)$ in unified semantic representation space. Beforehand, assuming that for a sentence $\mathbf{s}$, there exists a set of semantically related entities $\{e_s\}$. 
Afterwards, the semantic decoding vector $\hat{\mathbf{h}}$ can be obtained after passing $\mathbf{s}$ through the channel decoder at the receiver side. Accordingly, a unified semantic representation vector $\mathbf{v}_h$ can be determined.

Furthermore, some entity $e_+$ can be found from the semantically related entities and mapped to $\mathbf{v}_{e_+}$ as a positive sample. Meanwhile, $K$ irrelevant entities are randomly selected from the entity set $\{e\}$ in the knowledge base and the corresponding unified semantic representation vectors $\{\mathbf{v}_{e_-}\}$ will be regarded as negative samples where $K$ is for controlling the ratio of positive and negative samples.
On this basis, we take an InfoNCE loss function \cite{he2020momentum}, which can be expressed as Eq.~\eqref{eq_infonce} on Page \pageref{eq_infonce}.
\begin{figure*}
    \begin{equation}
\mathcal{L}_\text{InfoNCE} =
- \log{\left(\frac
{\exp(\mathbf{v}_h \cdot \mathbf{v}_{e_+} / \tau)}
{\exp(\mathbf{v}_h \cdot \mathbf{v}_{e_+} / \tau) + \sum_{e_k \in \{e_-\}}\exp(\mathbf{v}_h \cdot \mathbf{v}_{e_k} / \tau)}
\right)}.
\label{eq_infonce}
\end{equation}
    \hrulefill
\end{figure*}
Here $\tau$ is the temperature hyperparameter. Finally, through back propagation, the parameters of each layer in NNs will be updated. The overall process is shown in Algorithm ~\ref{alg-contrastive-training}.

\begin{algorithm}[t]
\caption{Training Process of the Unified Semantic Representation Based On Contrastive Learning.}
\label{alg-contrastive-training}
\begin{algorithmic}[1]
\renewcommand{\algorithmicrequire}{\textbf{Initialization:}}
\REQUIRE The parameters of the unified semantic representation space $\mathcal{U}$; the number of negative samples $K$. 
\renewcommand{\algorithmicrequire}{\textbf{Input:}}
\REQUIRE The input sentence $\mathbf{s}$ and the entity set $\{e_s\}$.  %semantically related triples $\{(e_p, r_{pq}, e_q)\}$
\renewcommand{\algorithmicrequire}{\textbf{Output:}}
\REQUIRE Embedding process of the entity $\mathcal{F}_{e}(\cdot)$ and mapping function of the channel decoded vector $\mathcal{F}_{\hat{h}}(\cdot)$.
\STATE Transmit the sentence $\mathbf{s}$ through the SemCom system, and obtain the channel decoded vector $\hat{\mathbf{h}}$ at the receiver side.
\STATE Take an entity $e_+$ from the entity set $\{e_s\}$ and calculate its embedding vector $\mathbf{v}_{e_+}$ by Eq.~\eqref{eq_e_mapping}.
\STATE Calculate the unified semantic representation vector of the channel decoding vector $\mathbf{v}_h$ by Eq.~\eqref{eq_s_mapping}.
\STATE Generate embedding vectors $\mathbf{v}_{e_-}$ as negative samples based on $K$ randomly selected irrelevant entities ${\{e_-}\}$ belonging to the entity set $\{e\}$ in the knowledge base.
\STATE Calculate the InfoNCE loss function by Eq.~\eqref{eq_infonce}.
\STATE Use stochastic gradient descent to optimize $\mathcal{F}_{e}(\cdot)$ and $\mathcal{F}_{\hat{h}}(\cdot)$.
\end{algorithmic}
\end{algorithm}

\subsection{Simulation Results} 
\label{sc-results-dynamic}
\paragraph{Dataset and Parameter Settings}
\label{paragraph-dataset-dynamic}
Consistent with the settings in Section \ref{sc-results}, this section utilizes the preprocessed WebNLG v3.0 dataset \cite{gardent2017creating} and leverages the Transformer-based model as the underlying SemCom system. In particular, the architecture comprises a $3$-layer codec with a model dimension of $128$. The simulation is conducted on the AWGN channel.

During the simulation, the Transformer-based JSCC is first trained on plain text without the assistance of any knowledge. Based on the pre-trained Transformer, the proposed knowledge enhancement module with an evolving KG is subsequently trained. During the training process, the hyperparameters are carefully calibrated for better performance. Specifically, the number of negative samples, denoted as $K$, is fixed at $63$, and the temperature parameter $\tau$ is set to $0.2$. Moreover, the distance threshold $\lambda$ is fixed at $1.16$.

\begin{figure}[t]
\centering
\includegraphics[width=0.4\textwidth]{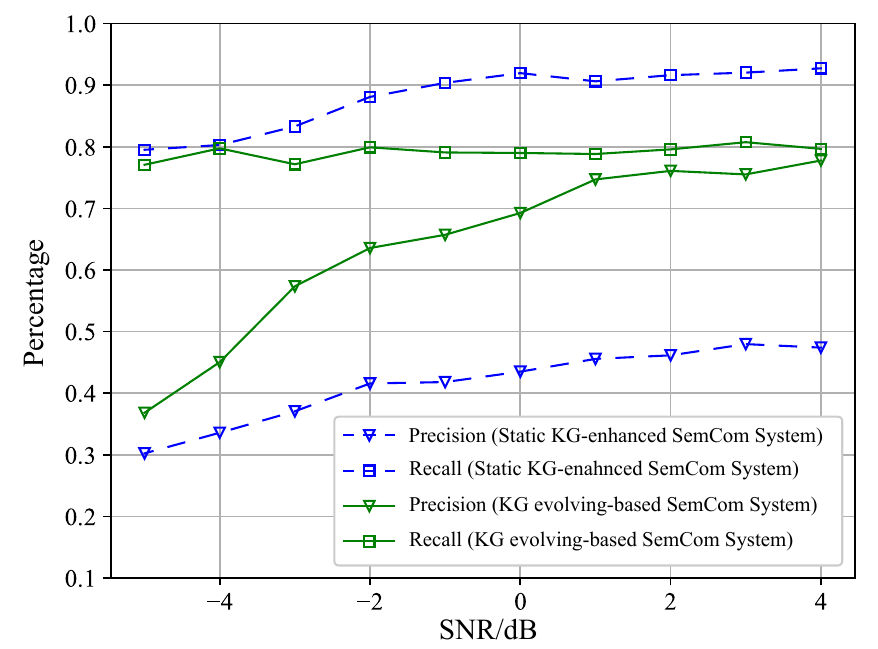}
\vspace{-0.2cm}
\caption{The comparison of precision and recall rate between the newly developed KG evolving-based approach and the static KG-based one in Section \ref{sc-KG-semcom}.}
\label{fig-dynamic-KG-decoder-results-comparison}
\end{figure}

\begin{figure*}[t]
\centering
\includegraphics[width=0.8\textwidth]{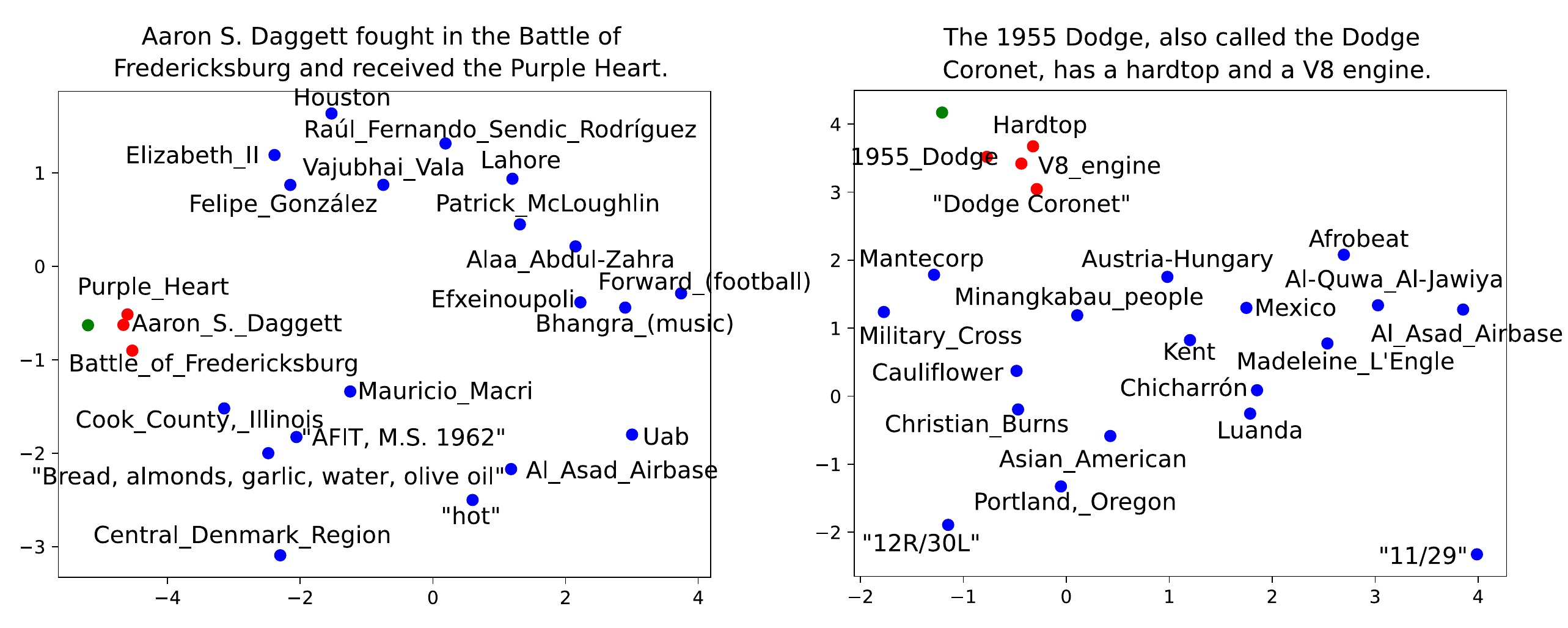}
\vspace{-0.2cm}
\caption{Examples of spatial visualization of unified semantic representation based on t-SNE dimensionality reduction algorithm \cite{van2008visualizing}.}
\label{fig-dynamic-KG-decoder-results-visualization}
\end{figure*}

\paragraph{Numerical Results}
\label{paragraph-results-dynamic}
The performance of the newly developed KG evolving-based module versus SNR is experimentally validated, with the corresponding precision and recall rate depicted in Fig.~\ref{fig-dynamic-KG-decoder-results-comparison}. Notably, the static KG-enhanced approach discussed in Section \ref{paragraph-results} is compared as well. Fig.~\ref{fig-dynamic-KG-decoder-results-comparison} manifests that the recall rate of the proposed KG evolving-based approach remains consistently stable at around $80\%$, while demonstrating a marginal decrease compared to the method detailed in Section \ref{paragraph-results}. On the contrary, the precision of the KG evolving-based approach exhibits a discernible superiority (i.e., $70\%$ at $0$ dB) and achieves about $80\%$ at higher SNRs. This outcome holds substantial positive implications, as a notable improvement in precision serves to mitigate the potentially misleading effects on the decoder stemming from missing or inaccurate knowledge.

To visualize the impact of unified semantic representations, a t-SNE dimensionality reduction algorithm \cite{van2008visualizing} is leveraged to project the vectors onto a $2$-dimensional space, with the corresponding results presented in Fig.~\ref{fig-dynamic-KG-decoder-results-visualization}. Notably, the red dots indicate the embedding results of related entities, while the blue dots represent those of irrelevant entities. Furthermore, the embedding of the semantic decoding vector is marked by a green dot. Therefore, it validates that the contrastive learning-based training method successfully maps semantically related entities to the vicinity of the semantic decoding vector, thereby enabling the feasibility of a distance-based extraction algorithm.

\begin{figure*}[t]
\centering
\includegraphics[width=0.8\textwidth]{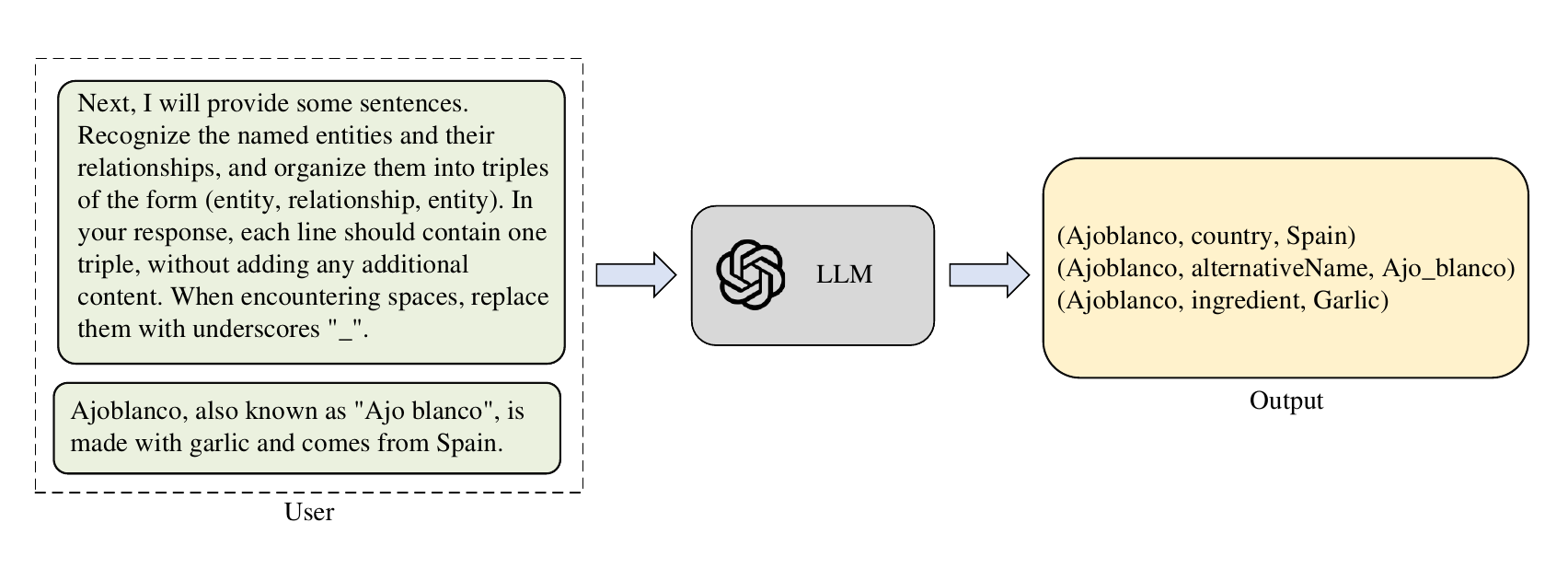}
\vspace{-0.5cm}
\caption{An example of LLM-assisted data augmentation solution for knowledge acquisition.}
\label{llm-solution}
\end{figure*}

\section{LLM-Assisted Data Augmentation for the KG Evolving-based SemCom System}
\label{sc-llm}
After discussion on the KG evolving-based SemCom system, whose semantic representation is unified by contrastive learning, we shed some light on LLM-assisted data augmentation solutions for knowledge acquisition. 

\subsection{Descrption of the LLM-Assisted Data Augmentation solution}
The KG evolving-based approach proposed in Section \ref{sc-KG-semcom-dynamic} can adaptively update the recorded factual statements in the knowledge bases. However, within SemCom, the un-annotated transmitted content proves challenging to directly correlate with existing knowledge bases. As a world model \cite{radford2018improving}, LLMs can competently fill the deficiency and be utilized for data augmentation by extracting the knowledge without the apriority of the transmitted content. In other words, by designing appropriate prompts, it is feasible to instruct LLMs to incorporate relevant prior knowledge or constraint conditions. More importantly, by embracing a zero-shot learning paradigm, LLMs can extract knowledge without manual annotation of relevant datasets or retraining of models, thereby enhancing system flexibility.

Fig.~\ref{llm-solution} illustrates the envisioned integration of LLMs to augment SemCom in Section \ref{sc-KG-semcom-dynamic}. It can be observed Fig.~\ref{llm-solution} that by using appropriate prompts, LLMs can identify named entities and predict relationships, so as to more swiftly formulate factual triples. Hence, this LLM-assisted data augmentation further fortifies the system's capability to manage dynamic knowledge effectively.

\subsection{Simulation Results}
\paragraph{Dataset}
The feasibility of the LLM-assisted SemCom scheme discussed in this section is also validated through simulation experiments. Specifically, the experiment simulates scenarios with incomplete knowledge matching by intentionally omitting a subset of the original knowledge dataset. To address this, a solution based on PE is employed to leverage the LLM for generating relevant knowledge, which is then integrated into the existing knowledge base. In detail, the experiment utilizes the GPT-3.5 Turbo model \cite{peng2023gpt}, incorporating $1000$ samples from the original dataset. The data generated by the expansive model is utilized for an overall system performance simulation, evaluating its efficacy in SemCom tasks. Among the $1000$ samples produced by the LLM, $800$ of them are divided into the training dataset for instructional purposes, while the remaining $200$ samples constitute the test dataset.

\begin{figure*}[t]
\centering
\includegraphics[width=0.8\textwidth]{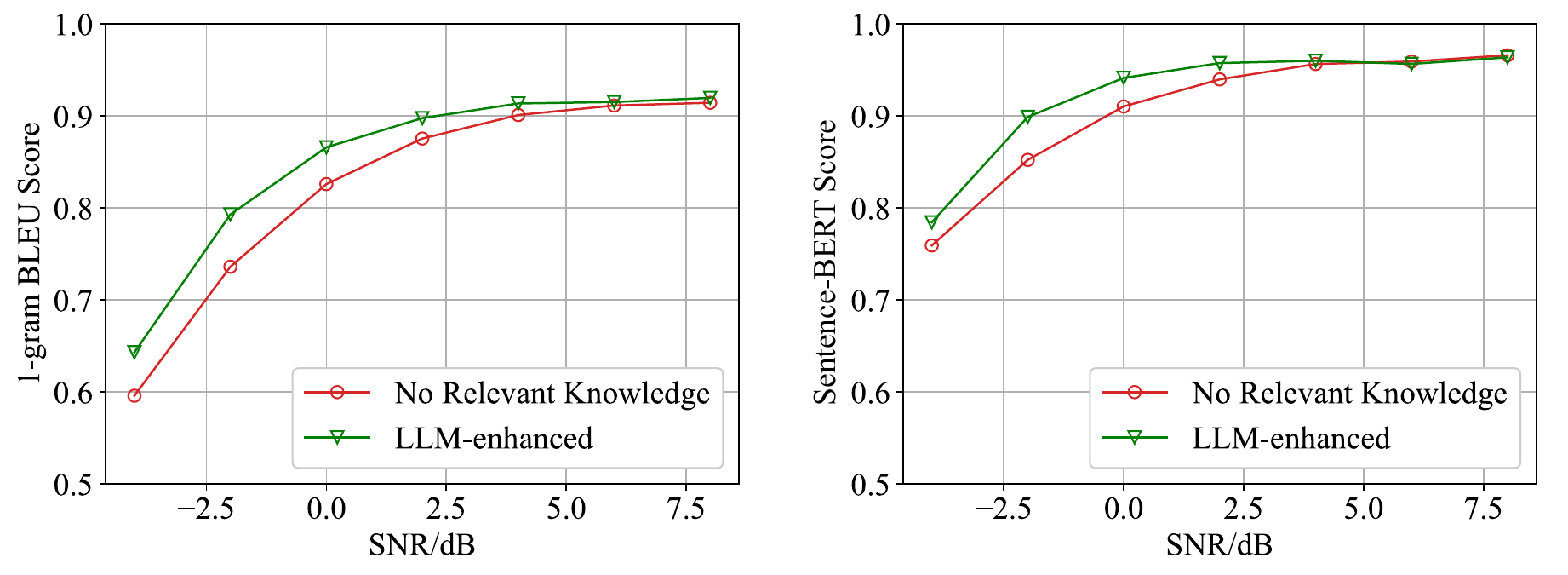}
\vspace{-0.5cm}
\caption{The performance evaluation of the proposed LLM-enhanced system.}
\label{llm-results}
\end{figure*}

\paragraph{Numerical Results}
Fig.~\ref{llm-results} illustrates the system's performance on the test dataset generated by the LLM. It can be observed from Fig.~\ref{llm-results} that the LLM possesses the ability to compensate for missing knowledge to a considerable degree, thus promising to improve the overall performance of the system. Particularly in scenarios with lower SNRs, the knowledge generated by the LLM contributes an approximate increase of $0.05$ in the BLEU score, while a similar phenomenon is observed for the Sentence-BERT score. Therefore, it validates the efficacy of the LLM-empowered scheme, and could further bolster SemCom's capability in processing and comprehending knowledge. 
%, closely resembling the results obtained with the original dataset. 

\section{Conclusion}
\label{sc-conclusion}
This chapter aims to investigate feasible and effective approaches for interplaying KGs and SemCom. It begins with an in-depth introduction to knowledge-processing challenges within SemCom, and proceeds to thoroughly investigate the limitations of pertinent existing studies. Subsequently, a KG-enhanced SemCom system is proposed, in which a knowledge extractor is leveraged to effectively exploit knowledge from KGs and facilitate semantic reasoning and decoding at the receiver side. On this basis, a contrastive learning-based optimization strategy for evolving KGs is described. Additionally, this chapter discusses methods for data augmentation through the utilization of LLMs. Extensive simulation results demonstrate that the proposed KG-enhanced SemCom systems can benefit from the prior knowledge in the knowledge base with significant performance gains. Looking ahead, future research could focus on further harnessing the semantic understanding and generation capabilities of LLMs to manage complex contextual and semantic information.

\printbibliography
\end{document}